\documentclass[pdflatex,sn-standardnature]{sn-jnl}
\jyear{2025}
\usepackage{setspace}

\usepackage{array,multirow}
\usepackage[normalem]{ulem}
\usepackage{caption}
\usepackage{subcaption}
\usepackage{pifont}

\usepackage{booktabs}
\usepackage{verbatim} 
\usepackage{cleveref}
\usepackage{hyperref}
\usepackage{float}
\usepackage{xfrac}
\usepackage{tabularx}
\usepackage{float}

\begin{document}

\title{\textbf{Explainable AI: Learning from the Learners}}

\author[1]{\fnm{Ricardo} \sur{Vinuesa}}\email{rvinuesa@umich.edu}

\author[2]{\fnm{Steven L.} \sur{Brunton}}\email{sbrunton@uw.edu}

\author[3]{\fnm{Gianmarco} \sur{Mengaldo}}\email{mpegim@nus.edu.sg}

\affil[1]{\orgdiv{Department of Aerospace Engineering}, \orgname{University of Michigan}, \orgaddress{\street{Ann Arbor, Michigan 48109}, \country{USA}}, Email: \url{rvinuesa@umich.edu}}

\affil[2]{\orgdiv{Department of Mechanical Engineering}, \orgname{University of Washington}, \orgaddress{\street{Seattle, Washington 98195}, \country{USA}}, Email: \url{sbrunton@uw.edu}}

\affil[3]{\orgdiv{National University of Singapore}, \orgname{National University of Singapore}, \orgaddress{\street{Engineering Drive 1, Singapore, 117575}, \country{Singapore}}, Email: \url{mpegim@nus.edu.sg}}

\maketitle




\section*{Abstract}

Artificial intelligence now outperforms humans in several scientific and engineering tasks, yet its internal representations often remain opaque. In this Perspective, we argue that explainable artificial intelligence (XAI), combined with causal reasoning, enables {\it learning from the learners}. Focusing on discovery, optimization and certification, we show how the combination of foundation models and  explainability methods allows the extraction of causal mechanisms, guides robust design and control, and supports trust and accountability in high-stakes applications. We discuss challenges in faithfulness, generalization and usability of explanations, and propose XAI as a unifying framework for human-AI collaboration in science and engineering.

\section*{Introduction} \label{sec:intro}

Scientific progress has long depended on building interpretable models of complex systems, from Newton’s laws to the Navier--Stokes equations, which encode mechanistic understanding and enable prediction, control and design. In recent years, however, machine learning (ML) has begun to transform this paradigm. Deep neural networks can now infer relationships in high-dimensional data that are inaccessible to classical theory, often matching or surpassing human experts in tasks ranging from protein folding~\cite{jumper2021highly} to turbulence control~\cite{Guastoni2023_drl}. This new reality raises a profound question: if machines can learn representations of physical systems that outperform human models, can we, in turn, learn from the learners? Addressing this question requires explainable artificial intelligence (XAI)~\cite{lundberg2017}: a suite of methods that make the internal logic of ML systems interpretable to human reasoning.

Explanations have always been central to scientific inquiry: they connect abstract models to causal understanding and experimental validation. In modern deep learning, however, the representations learned by a model are rarely transparent~\cite{rudin2019stop}. XAI thus provides the missing interface between machine predictions and human understanding, allowing researchers to inspect which features drive the decisions of a model, quantify their causal influence and assess whether the learned mechanisms are physically meaningful.

The connection between explainability and causality~\cite{campsvalls2023discovering,carloni,Cremades2025} is particularly relevant for science and engineering. Causality formalizes the notion of intervention, and is thus essential to scientific reasoning. Recent developments in causal inference and information theory allow decomposition of causal effects into unique, redundant, and synergistic components~\cite{martinez-sanchez2024surd}, enabling a more granular view of how different variables contribute to an outcome. Such tools help distinguish true physical mechanisms from spurious correlations that may arise in complex datasets. When combined with XAI techniques such as SHapley Additive exPlanations (SHAP)~\cite{lundberg2017} or integrated gradients~\cite{sundararajan2017axiomatic}, they enable a mechanistic understanding of deep-learning models: which regions of a flow field, which frequencies in a spectrum, or which molecular configurations are most responsible for a predicted outcome. 
This causal-explanatory synergy opens new opportunities to extract scientific laws from data-driven systems.

A key point in this context is generalizability. Models that are fully explainable in a narrow domain but fail when extrapolated to new conditions provide limited scientific insight. Explainability therefore complements generalization: only by identifying the underlying mechanisms can one predict the behavior under unseen parameters. 
Explainability also plays a decisive role in identifying the mechanisms behind extreme events. 
These rare but high-impact occurrences (e.g. extreme weather or structural failures) are often underrepresented in the training datasets and are thus poorly captured by statistical models. Indeed, AI models tend to overfit frequent regimes while missing rare ones~\cite{wei2025xai4extremes}. 
XAI methods can reveal the subtle precursors leading to such events, differentiating causal chains from correlations. This capability is critical for safety and resilience in both natural and engineering systems, where predicting extremes may depend on uncovering weakly correlated but dynamically dominant features. This information, in turn, can help shape ML-model behavior, thereby improving forecasts of rare events.

Recent advances in self-supervised learning and diffusion models~\cite{yang2023diffusion} further expand the scope of explainability. These approaches learn by context rather than explicit supervision, discovering the statistical structure of their environment. By embedding physics into such architectures (through inductive biases, conservation constraints or equivariance principles) these models can capture the latent organization of complex systems. Yet, the more powerful these models become, the more opaque their internal reasoning tends to be. Interpretable training signals, causal priors and physically meaningful embeddings are therefore essential for ensuring that generalization arises from the identified mechanisms and not by coincidence.

It is also essential to note that explainability can be illusory. Not every attribution map or saliency pattern constitutes a true explanation. Post-hoc justifications (especially those mimicking human reasoning) can create a false sense of understanding. Anthropomorphic metaphors (e.g., describing networks as ``attending'' or ``reasoning'') may obscure what a model actually computes~\cite{ameisen2025circuittracing,mengaldo2024explain}. 
Thus, explainability itself must be subjected to rigorous validation~\cite{turbe2023evaluation}, with metrics for faithfulness and stability ensuring that explanations correspond to real causal influence~\cite{wei2024revisiting}.

\begin{figure}[h]
\includegraphics[width=\textwidth]{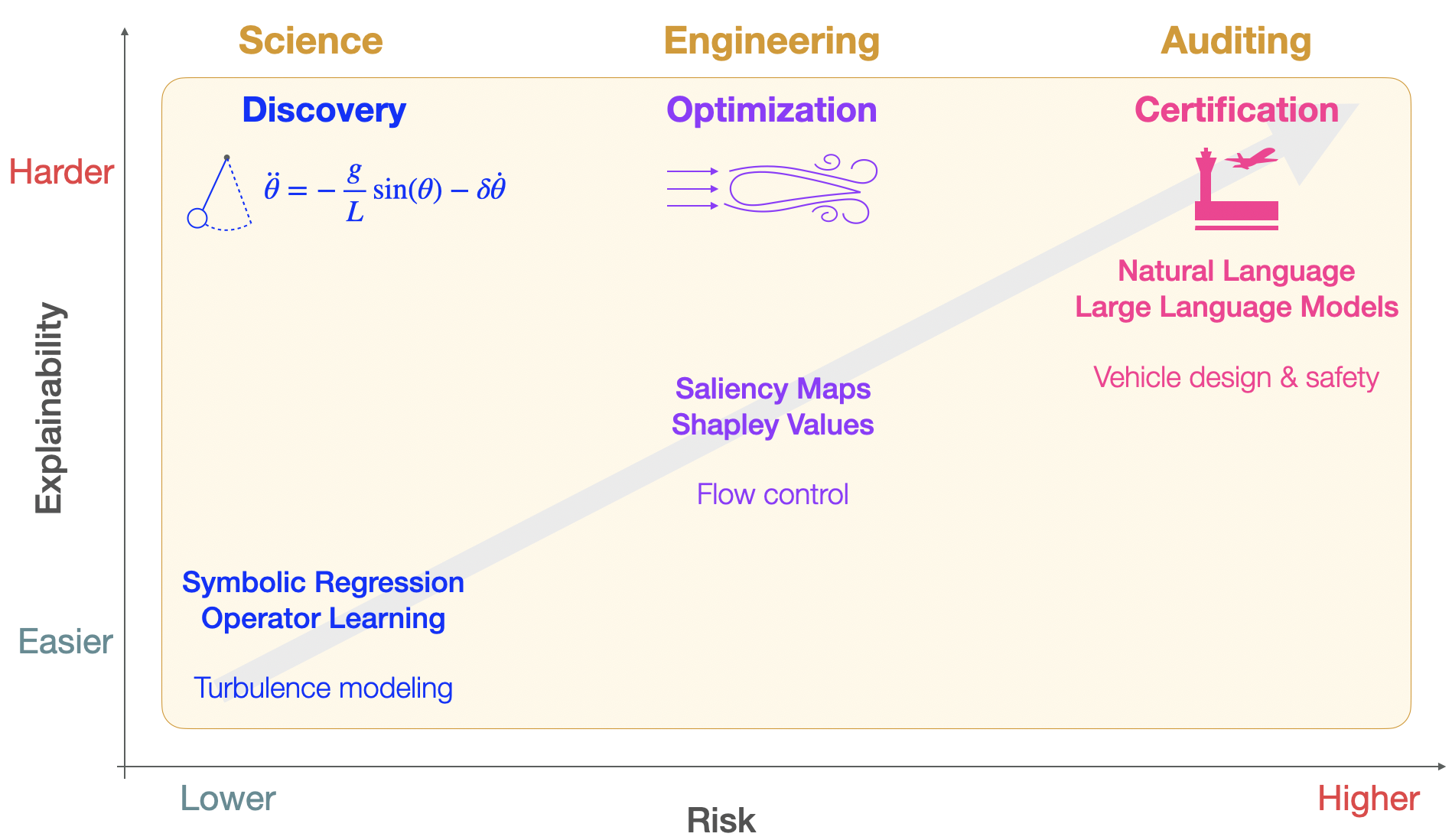}
\caption{{\bf Summary of the application areas for explainable AI discussed in this study.} The three main areas of discovery, optimization and certification (generally associated with science, engineering and auditing) are ranked in terms of the risk of the application and of how easy it is to obtain explanations in that context. For each area we list relevant methods (in bold face) and an example of application in the context of fluid mechanics for vehicle aerodynamic design.}
\label{fig:summary}
\end{figure}

In this Perspective, we argue that explainable AI offers not just interpretive insight but a framework for scientific collaboration between humans and machines. As summarized in Figure~\ref{fig:summary}, we focus on three domains: discovery, optimization and certification, which are connected with science, engineering and auditing. In discovery, XAI aids in unveiling causal mechanisms and uncovering governing equations from data. In optimization, it guides design and control by revealing which parameters, mechanisms or spatial regions drive performance. And in certification, it achieves trust and accountability, allowing automated systems to be audited and aligned with human values. These domains are tightly coupled: insights discovered through explainable models inform optimization strategies, and the interpretability required for certification constrains how discoveries are validated and deployed. Explainability thus acts as the link between data-driven learning, mechanistic understanding and ethical governance.

Deep learning has paradoxically expanded both the opacity and the potential of scientific models. While black-box architectures outperform interpretable surrogates, they also contain within them rich representations of physical phenomena. If extracted properly~\cite{Brunton2016pnas}, these representations may reveal new principles (similarly to how symbolic-regression methods~\cite{cranmer2023interpretable} distilled empirical laws from data), but now at a scale spanning billions of parameters and multidimensional manifolds~\cite{vinuesa2024decoding}. Explainable AI provides the tools for this extraction, translating machine-learned representations into human knowledge. As the boundaries between simulation, optimization and automated reasoning blur, the ability to interpret how machines learn will determine whether artificial intelligence becomes a scientific collaborator or merely a computational tool.

Ultimately, learning from the learners requires more than decoding predictions: it involves understanding the underlying reasoning processes. Through linking causal inference, interpretability and generalization, XAI can turn data-driven models into engines of discovery and control that not only predict the world but also explain it. The following sections explore how this vision unfolds across the three core domains, each bringing distinct challenges and converging towards a shared goal: a new form of science in which humans and machines co-create understanding.

\section*{Explainable AI for Discovery}
There is tremendous potential for AI to drive progress in scientific discovery, with examples already emerging in fields as diverse as molecular biology~\cite{Noe2020ARPC,jumper2021highly,baek2022deep}, fusion~\cite{degrave2022magnetic}, astrophysics~\cite{tamayo2020predicting,parker2024astroclip,angeloudi2024multimodal}, and fluid dynamics~\cite{Ling2016jfm,Loiseau2017jfm,fukami2024data}. 
It is likely that the future of science will involve an even deeper collaboration between humans and AI agents.  
However, many of our most powerful deep-learning architectures are opaque and do not readily provide insights into how outcomes and predictions are generated.  
This opacity is in direct conflict with the scientific method~\cite{mengaldo2024explain}, which demands detailed explanations and causal mechanisms, which provide the ability to justify conclusions, verify results, and design new hypotheses to be tested.  
If we aim to remain active participants in this collaboration, we must not only seek models with the highest prediction accuracy, but instead design models that uncover simple, explainable causal mechanisms.   

Fortunately, there are several techniques that help provide the explainability required for AI agents to work collaboratively with human scientists.  
Explainability in a scientific context typically relies on identifying causal mechanisms that are \emph{parsimonious}, meaning that they are as simple as required to describe the data, and cannot be made simpler.
Parsimonious models tend to be more interpretable by design, and more generalizable, as they discourage overfitting and distill the key underlying mechanisms of a system. 
This principle of parsimony, also known as Occam's razor, has been the gold standard in physics for 2,000 years, from Aristotle to Einstein, and it is unlikely that this paradigm will be upended in a few decades of AI research.  

Of course, we hope that AI agents will uncover correlations and mechanisms that are more sophisticated than humans are able to explore directly.  
However, complexity is not incompatible with explainability, provided the explanations are \emph{composable} in terms of simpler concepts.  
Many complex systems are understood through sequences of individually explainable steps, even when the overall system requires substantial expertise to master, such as operating a nuclear reactor or assembling an aircraft.
In scientific discovery, we may therefore accept a reduction in descriptive accuracy in exchange for models that expose fundamental and generalizable mechanisms.
For example, Kepler’s elliptical orbits were less accurate than Ptolemy’s astronomical system, yet they revealed a simpler and more generalizable structure that ultimately enabled further scientific progress. 
Similarly, Galileo's constant gravity model was less accurate in explaining the trajectories of falling objects that Aristotle's density-based approach, although it captured a fundamental underlying principle.  Indeed, after this explainable bit of physics, it was then possible to improve the model with additional physics, such as wind resistance, and vortex shedding, among others. 

There are several techniques at our disposal to promote parsimonious, and hence explainable, AI models.  
Sparsity and low-dimensionality have long been used in machine learning to constrain models toward representations that isolate the essential degrees of freedom of a system.
By limiting the number of active variables or latent dimensions, these approaches support interpretability even when the underlying implementations are complex.
Sparse symbolic regression provides a particularly clear path towards parsimonious models, as the explicit objective is to discover simple equations that describe the data or the function learned by a model.


Symbolic regression is a particularly useful technique in machine learning for scientific discovery, resulting in models that take the form of parsimonious symbolic expressions.  
These models are often more interpretable to humans, as they can uncover the causal relationships necessary for explainability. 
Today there are two main approaches to symbolic regression: genetic programming to grow function trees compositionally, and sparse regression in a library of candidate terms.  
These approaches are complementary, operating in different regimes of expressivity, data efficiency, and computational cost~\cite{brunton2025machine}.  In the following, we will briefly review each approach and highlight how they have been useful in explainable AI-enabled scientific discovery.  

Genetic programming (GP) grows increasingly sophisticated function expressions through composition of simpler basis functions, such as addition, multiplication, trigonometric functions, and other elementary operations.  
Although GP was widely used for decades, it became a powerful tool for parsimonious scientific discovery in the work of Bongaard and Lipson~\cite{Bongard2007pnas} and Schmidt and Lipson~\cite{Schmidt2009science}.  
A key contribution of this work was their emphasis on the importance of parsimony in symbolic learning, using the Pareto frontier of model complexity versus accuracy to identify expressions that are both simple and descriptive.  
In the intervening years, great strides have been made~\cite{petersen2019deep,vaddireddy2019equation,udrescu2020ai,cranmer2023interpretable}, with PySR~\cite{cranmer2023interpretable} emerging as a powerful general-purpose tool for symbolic regression in scientific discovery.  

More recently, sparse regression has emerged as a particularly effective technique for parsimonious symbolic regression in scientific discovery~\cite{Brunton2016pnas}. 
The sparse identification of nonlinear dynamics (SINDy) algorithm identifies the fewest terms in a library of candidates functions needed to describe a given system, typically the right-hand side of a differential equation.  
By construction, these models have relatively few parameters, making them data efficient and contributing directly to their interpretability and generalizability. 
SINDy is particularly extensible because it is based on generalized linear regression in a library of nonlinear functions, with variants that incorporate actuation and control~\cite{zolman2024sindy}.  
Importantly for scientific discovery, it is possible to enforce symmetries and constraints derived from known physics~\cite{Loiseau2017jfm}, such as energy conservation in incompressible fluids, which in turn makes it possible to guarantee model stability~\cite{kaptanoglu2021promoting}.  
The dual perspective is to learn what symmetries are present in a dataset or model, providing an additional and powerful form of physical explainability~\cite{otto2023unified}. 

One of the most important applications of sparse symbolic regression for scientific discovery is its extension to partial differential equations (PDEs)~\cite{Rudy2017sciadv, Schaeffer2017prsa}. 
By augmenting the library with partial derivatives, it is possible to learn a sparse representation of the governing PDE directly from data. 
While this approach is generally sensitive to noise, it has been augmented with the weak formulation~\cite{messenger2021weak,reinbold2021robust} and ensemble learning~\cite{fasel2022ensemble} to dramatically improve noise robustness.  
These advances have enabled the discovery of entirely new, interpetable physics models in domains including closure modeling for turbulent fluids~\cite{beetham2020formulating,beetham2021sparse,schmelzer2020discovery}, plasmas~\cite{alves2020data}, Galerkin models~\cite{Loiseau2017jfm}, active matter~\cite{supekar2023learning}, and geophysical flows~\cite{zanna2020data}. 
Other symbolic approaches for PDE learning have also shown promise~\cite{long2019pde}.

Closely related to discovering parsimonious models of system dynamics is the importance of identifying effective coordinate systems in which to represent the dynamics.  
Many major scientific breakthroughs were preceded by the discovery of appropriate coordinate; for example, the Copernican model, which placed the sun at the center of our solar system, directly enabled the discovery of Kepler's laws of planetary motion.  
One of the great strengths of modern AI systems is their ability to digest vast quantities of data and uncover dominant, actionable patterns, making them well suited for discovering such coordinates. 
This has led to one of the most important tools for explainable AI: the autoencoder (AE)~\cite{kingma2013auto,Lusch2018natcomm,Champion2019pnas,lee2020model,conti2023reduced,mounayer2024rank}.  
Autoencoders operate by reconstructing input data while imposing an information bottleneck, typically through a low-dimensional latent space, forcing high-dimensional data to be distilled into a small number of coordinates.  
In this sense, AEs may be viewed as a nonlinear generalization of the singular value decomposition (SVD), which is itself a data-driven generalization of the Fourier transform~\cite{Brunton2022book}.  
Autoencoders have therefore been widely used to improve explainability in scientific machine learning, for example to promote sparse models in SINDy~\cite{Champion2019pnas,bakarji2023discovering}, or by identifying coordinates that enable even more stringent assumptions, such as strictly linear latent dynamics~\cite{Lusch2018natcomm,Takeishi2017nips,Yeung2017arxiv,Wehmeyer2018jcp,Mardt2018natcomm,Otto2019siads}. 
Even when the network implementation is complex, the explicit identification of a low-dimensional latent space provides a meaningful and interpretable representation.  
Further, Cranmer et al.~\cite{cranmer2021disentangled} introduced the notion of \emph{disentangled sparsity} to derive symbolic expressions for each of the latent variables in terms of a sparse subset of the high-dimensional input variables.  

A complementary form of explainability in AI-enabled scientific discovery arises from restricting models to represent specific physical functions or mechanisms.  
We have already seen this in the example of turbulence closure modeling, where the focus is put on modeling a specific physical mechanism, such as the Reynolds stresses~\cite{Ling2016jfm}. 
Regardless of the internal complexity of the model, if its purpose is tightly constrained to a well-defined physical concept, then explainability is embedded in the function that the model is designed to represent.  
This perspective has been particularly influential in computational chemistry~\cite{Noe2019science,Noe2020ARPC}, for example through the use of free energy and related concepts to impose physical meaning on deep learning models~\cite{Noe2019science}.  
Related examples include Lagrangian and Hamiltonian neural networks~\cite{greydanus2019hamiltonian,cranmer2020lagrangian}, which leverage deep learning to represent fundamental physical functions with clear interpretations.  
Incorporating symmetries into learning architectures is also used to improve generalization and embed deeply intuitive physical principles into otherwise opaque models~\cite{Loiseau2017jfm,miller2020relevance,batzner20223,brandstetter2022clifford,brandstetter2022lie,otto2023unified}. 
More broadly, scientific explainability has often been bootstrapped by identifying and naming recurring functional structures.  For example, the Bessel functions are defined by recursion relations to solve classes of differential equations that occur frequently in science and engineering, providing a compact and interpretable abstraction for otherwise complex behavior.  

\begin{figure}[t]
    \begin{center}
        \includegraphics[width=\textwidth]{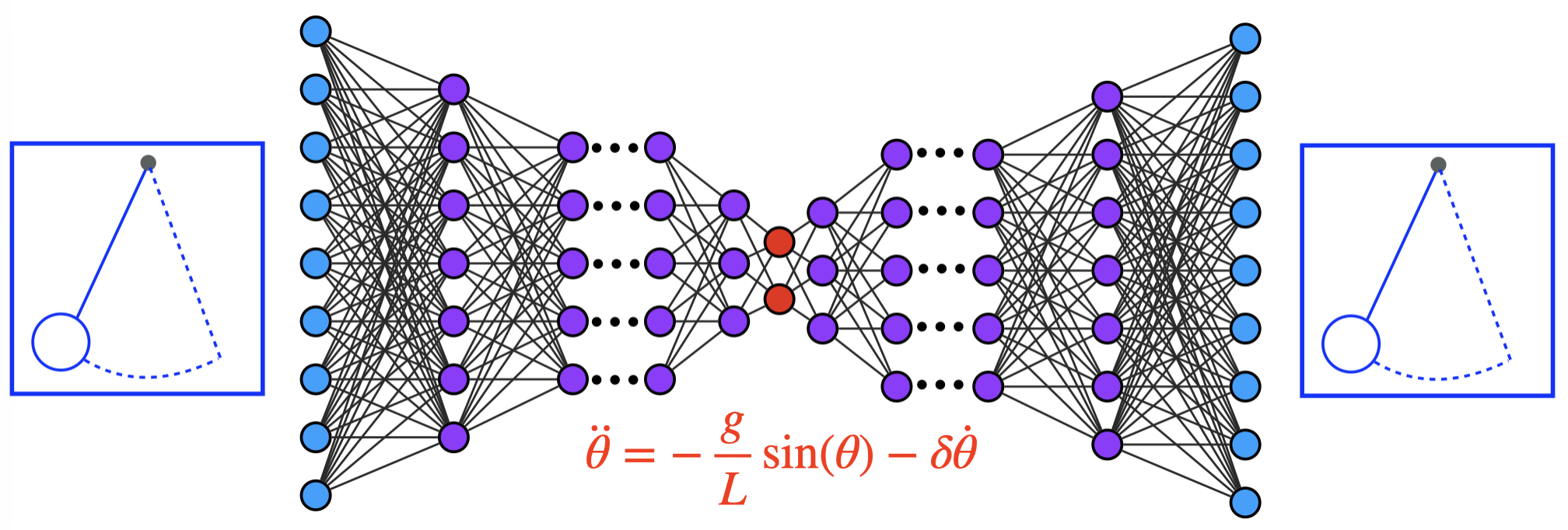}
        \caption{\textbf{Autoencoders learn coordinates for parsimonious dynamics.}  Discovering an effective coordinate system and a parsimonious description for mechanistic behavior is a goal in XAI for scientific discovery. In this schematic, a video of a swinging pendulum is the input to the autoencoder, which parameterizes the kinematics with two latent variables, $\theta$ and $\dot{\theta}$.  It is then possible to learn parsimonious models of the dynamics in this latent space, for example using SINDy~\cite{Champion2019pnas}.}\label{Fig:XAI_Discovery}
    \end{center}
\end{figure}


\section*{Explainable AI for Optimization} \label{sec:optimization}

Optimization problems in science and engineering increasingly rely on machine-learning (ML) models with predictive~\cite{Du2024CoNFiLD,solera2024beta} and control~\cite{mnih2015human,lillicrap2016continuous} capabilities exceeding those of traditional approaches. However, exploiting these models for design or real-time control requires understanding why they make certain decisions and which variables or regions of the system drive those outcomes. Explainable artificial intelligence (XAI)~\cite{lundberg2017,erion2021} therefore plays a central role in obtaining this type of insight, enabling both improved performance and deeper physical understanding. Classical optimization methods typically rely on gradient-based sensitivity analysis~\cite{martins2013multidisciplinary} or adjoint methods~\cite{jameson1988adjoint} to identify influential parameters. In contrast, modern ML models (especially deep neural networks) require explainability techniques adapted to high-dimensional regression and classification tasks. The SHapley Additive exPlanation (SHAP)~\cite{shapley1953} framework, with its theoretical foundation in cooperative game theory, has become a cornerstone of post-hoc attribution methods for complex models. Kernel SHAP~\cite{lundberg2017} provides a flexible (although computationally expensive) approximation, while gradient-based SHAP~\cite{erion2021} offers scalability by exploiting backpropagation. Other relevant XAI methods are integrated gradients~\cite{sundararajan2017axiomatic} and saliency maps~\cite{simonyan2014deep}. In the context of engineering optimization, these methods reveal which input features most strongly influence the optimization objective, from aerodynamic performance to energy efficiency. They have been successfully applied across fluid mechanics~\cite{Cremades2024,Cremades2025}, heat transfer~\cite{Larranaga2024}, combustion~\cite{Zhao2024MachineCFDHybrid} and renewable-energy applications~\cite{Singh2022ANN_CylindricalRibs}, among others. Such analyses clarify the physical mechanisms learned by the model and highlight potential strategies for improving designs and controlling systems beyond what the model itself proposes.

A recurring challenge is that high-performing ML models may learn representations that are highly efficient for the machine, yet unintuitive for humans. For example, deep reinforcement learning~\cite{Guastoni2023_drl,verma2018efficient} or diffusion-based design models~\cite{Vishwasrao2025DiffSPORT} often solve a control or optimization task in latent spaces that compress the dynamics into low-dimensional manifolds. While such manifolds are computationally advantageous, they obscure the physical mechanisms underlying the solution. Here, explainability offers a mechanism for translating between the internal representation of the machine and the human-understandable physical space. Gradient-based SHAP~\cite{erion2021} is particularly valuable because it can be computed at extremely high resolution, scaling to hundreds of millions of spatial points in parallel. This makes it possible to identify causal importance in the physical flow domain~\cite{Cremades2025}, even when the underlying network operates in a latent space. Consequently, SHAP-based explanations can ``invert'' the abstraction introduced by representation learning, mapping the compressed reasoning of the machine back into interpretable physical structures, such as coherent vortical regions, thermal layers, reactive fronts or stress concentrations~\cite{Cremades2024,Cremades2025}. This dual-space interpretability is essential in engineering, where optimization outcomes must ultimately translate into physical actions: changes to geometry, actuation or boundary conditions. Connecting latent-space decision making with physical-space mechanisms enables domain experts to validate, generalize and build upon machine-generated solutions. An example of this is the usage of gradient SHAP to guide flow control based on deep reinforcement learning: essentially, SHAP determines the most important structures in a turbulent flow, and using them to guide the DRL reward it is possible to achieve a very effective flow-control strategy~\cite{Beneitez2025ImprovingTurbulenceControl}. In fact, taking the example of drag reduction, it has been shown~\cite{Beneitez2025ImprovingTurbulenceControl} that using drag minimization as the reward is less effective than minimizing the physical mechanisms responsible for the drag, as the latter tackles the root rather than the symptom. This principle has been extended to other optimization problems~\cite{HeatonFung2023}, with key implications in science~\cite{choquet} and engineering~\cite{Shen2022ExplainerGeometricKnowledge}.

Optimization is inherently causal: improving a design requires knowing which intervention causes what effect. Recent developments in causality and information theory~\cite{shannon1948}, such as the SURD (synergistic, unique and redundant decomposition) framework~\cite{martinez-sanchez2024surd}, provide a richer decomposition of causal interactions. Note that computing SURD or related causal decompositions in physical space is prohibitively expensive for high-dimensional systems. To address this, causal inference is often performed in a latent space, where the dynamics are encoded as multivariate time series. Autoencoders~\cite{hinton2006reducing, lusch2018deep}, $\beta$-variational autoencoders ($\beta$-VAEs)~\cite{solera2024beta,vae,bvae} or other manifold-learning architectures naturally produce such representations. Within these manifolds, SURD can reveal the minimal causal structure governing the evolution of the system or the optimization objective. This causal structure then guides principled intervention strategies, for example identifying the specific latent directions that correspond to beneficial actuation paths, or the synergies between flow regions responsible for drag reduction. When combined with SHAP-based reconstructions in physical space, this creates a powerful causal-attribution framework: causality is identified in the latent manifold, then translated into physical mechanisms~\cite{arranz2024} that can be exploited e.g. for control.

Beyond attribution, gradient SHAP can act as a form of causal dimensionality reduction. Instead of compressing the system into a latent space, SHAP can reweigh the physical points of the domain according to their importance for a given task~\cite{Cremades2025}. This preserves interpretability and maintains direct correspondence with actuators, which naturally operate in physical coordinates. By pruning the domain to only the most important points one can build reduced-order models or controllers that focus exclusively on the dynamically relevant structures. This creates an alternative to latent-space modeling which is interpretable and physically grounded. Then, the SHAP-identified features may serve as a compressed but physically meaningful basis for reconstruction or closed-loop control~\cite{Beneitez2025ImprovingTurbulenceControl}. This physical-space dimensionality reduction is particularly appealing for engineering systems where actuation is spatially localized: SHAP highlights precisely those regions where interventions are most effective, enabling energy-efficient and robust optimization strategies. Thus, integrating explainability into optimization closes the loop between discovery and action: ML models propose an optimized solution or control policy, often identified in a latent space. Causal inference identifies how the solution emerges, pinpointing the interactions that drive improvements. Gradient-based SHAP represents these interactions in the physical space, producing interpretable attribution fields. Engineers can use these insights to design or refine actuation mechanisms, validate the reasoning of the model or generalize to new conditions. Then the system becomes safer, more transparent and more robust, particularly when optimization interacts with complex physics. This workflow applies to shape optimization, turbulent-flow control, structures, thermal management, reactive systems and multi-physics applications. It preserves the strengths of ML-driven optimization (speed, flexibility and nonlinearity) while ensuring physical interpretability.

\begin{figure}[H]
\centering
\includegraphics[width=\textwidth]{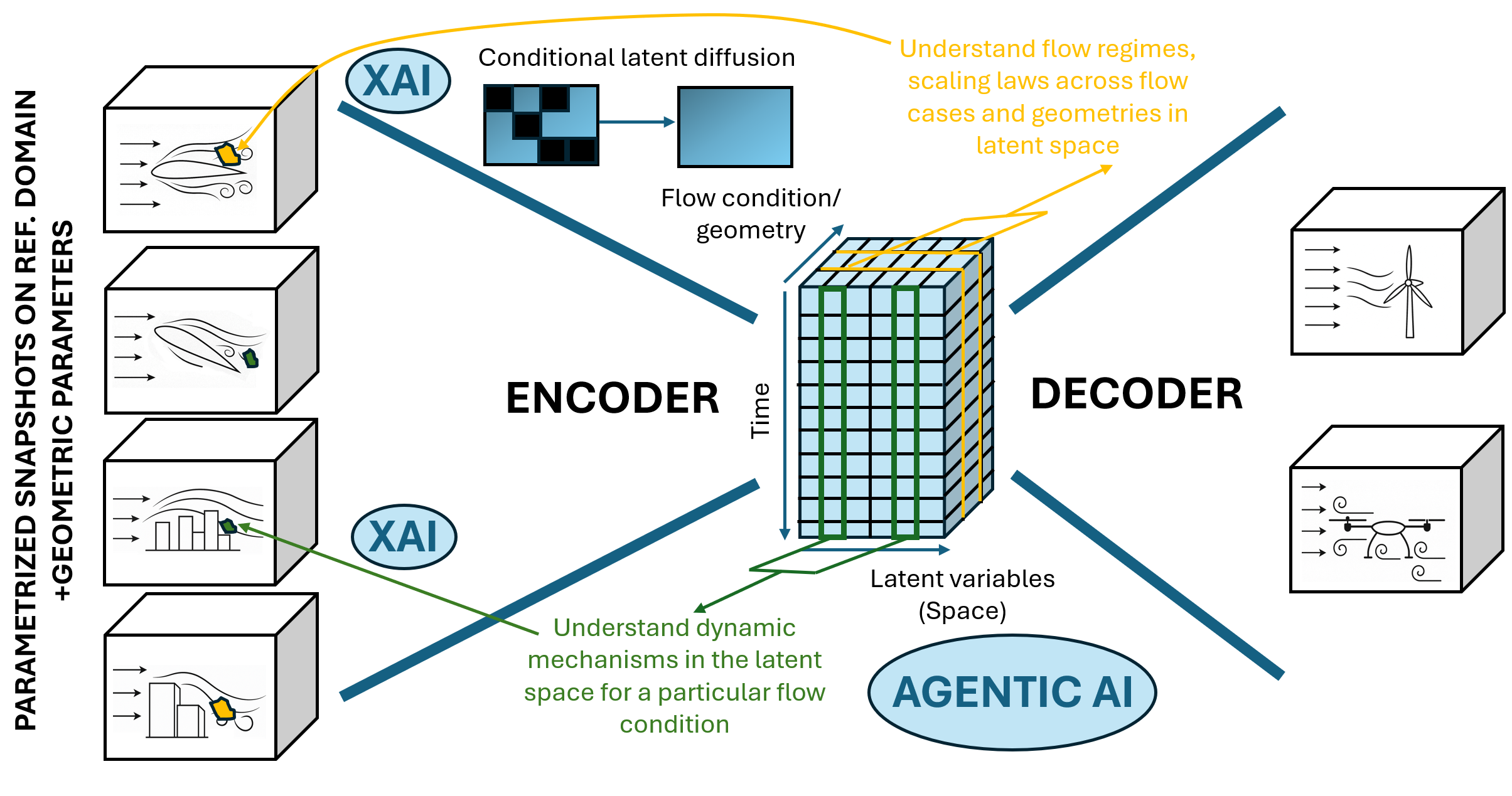}
\caption{{\bf Possible XAI-based framework for discovery and optimization.} We illustrate this framework with a fluid-mechanics problem. High-quality data from a range of geometries and flow conditions are used to train a foundation model~\cite{bommasani2022}, which can basically produce flow realizations in new geometries/conditions. The instantaneous fields for the various geometries are expressed on a common reference domain, and the geometry is defined based on a common parameterization, which is provided as an input to the encoder (potentially enabling generalizing to other geometries by properly choosing the parameters). The physical dimensions are encoded into disentangled latent variables via $\beta$-VAEs~\cite{solera2024beta}, and time is predicted by transformers~\cite{ref_easy}, such that each flow case maps into a 2D slice of the latent space. Then, each new flow condition is encoded into a new slice of the 3D latent space. Conditional latent diffusion~\cite{Du2024CoNFiLD} is used to produce new slices in the latent space for the various flow conditions, including unseen ones. When these get decoded, instantaneous realizations of those new conditions are obtained in physical space. Causal analysis~\cite{martinez-sanchez2024surd} can be used to understand physical phenomena for a particular flow case in the latent space, and in this case SHAP can connect physical regions with each of the latent variables~\cite{Cremades2024,Cremades2025}, thus providing physical understanding in an interpretable manner. An LLM-based agentic-AI system can be used to explore new phenomena by deciding which new flow conditions need to be investigated with the foundation model, thereby understanding the effect of various flow regimes, scaling laws and the impact of geometry. This will then be again mapped into physical space via SHAP, providing a complete framework for autonomous discovery very relevant for optimization in engineering and science.}
\label{fig:optimization}
\end{figure}

Explainable AI is emerging both as a diagnostic tool and a method for optimization and physical understanding. A possible framework for this is illustrated in Figure~\ref{fig:optimization}, exemplified by a fluid-mechanics problem: data from different flow cases (wings, cities...) and flow conditions (Reynolds number, Mach number, flow orientation...) are first compressed into a shared latent space via a $\beta$-VAEs~\cite{solera2024beta} (which encodes the spatial dimensions) and a transformer~\cite{ref_easy} (which perform the temporal predictions). Each flow condition is mapped into a slice of the latent space, and conditional latent diffusion~\cite{Vishwasrao2025DiffSPORT,Du2024CoNFiLD} is used to learn the distribution of the different slices of the latent space, potentially generalizing to new (unseen) conditions (wind turbines, drones...) as long as a good paramterization across geometries is found. This foundation model~\cite{bommasani2022} can be used as an engine for discovery by an agentic-AI system based on large language models (LLMs)~\cite{gottweis2025towards,vinuesa2026balancing}. Note that such foundation models have been proposed for discovery in chemistry~\cite{Wadell2025Foundation} and LLMs are being used for automated design optimization~\cite{Carreon2025Automated}. Such a system may use one slice of the latent space to understand dynamic mechanisms for a particular flow configuration based on causal analysis~\cite{martinez-sanchez2024surd}, and {\it decide} the next conditions to explore by defining new parameters and generating new latent spaces that can be decoded into a physical space, producing new relevant phenomena in that new configuration. SHAP can be key in this context, since it can connect regions of the physical space for certain system conditions with concrete latent representations. Thus, an agentic AI system which may reach scientific understanding for optimization in a latent space can be made interpretable in physical space through SHAP, which will enable connecting known physical phenomena with the physics newly discovered in the latent space. In summary, foundation models combined with XAI can help to reduce the complexity of a problem, enabling a more efficient optimization.

\section*{Explainable AI for Certification}

ML models are increasingly embedded into high-stakes scientific, engineering and social workflows, from climate prediction and natural-hazard early warning to structural health monitoring, aerospace design, offshore energy systems, autonomous robotics,  healthcare, and finance. 
As these AI-augmented systems transition from research prototypes to operational decision tools, a central question lies in how we certify that these models are trustworthy and safe for deployment. 
The response to this question has critical implications for liability: should the AI-augmented system fail and create significant damages to the affected entity, who should pay the price?

Traditional validation pipelines involve extensive testing prior to adoption and deployment. For instance, an aircraft would need to go through extensive ground-based testing of each component, and flight tests prior to entering into service. 
The on-board computers will also require extensive testing under various operating conditions to guarantee that fly-by-wire is safe and does not behave unexpectedly (failure in this realm can produce catastrophic consequences, e.g. Ref.~\cite{boeing-faa}).
Similarly, a medical doctor will need to undergo extensive specialized training before being able to practice and take decisions on a patient's well-being. 
When an aircraft fails or a medical doctor makes a wrong decision, the liability chain is relatively clear, albeit not always. This, because there are protocols in place to certify and audit what went wrong. In fact, many of these more traditional workflows tend to behave with ``deterministic causality'': a failure takes place and we can walk backwards to understand the steps that led to this failure. Obviously, this is not always the case: certain systems are inherently random (e.g., the financial markets). In these cases, a degree of failure is disclosed a-priori and is expected.
However, in engineering and healthcare applications such tolerance to failure is vastly reduced, and there are mechanisms in place to prevent catastrophic outcomes (or at least understand why they eventually happened). 

If we are now deploying AI-augmented solutions to some of these failure-adverse fields, such as engineering and medicine, what are the guardrails (if any) that we want to have in place to certify a given ML solution, and audit it in case of failure?

ML systems are statistical tools with a certain degree of randomness, whereby data distribution shifts and data points can cause them to fail, or at least not behave as expected. 
This could open a new frontier in scientific ML: the use of explainability as a pathway to formal certification and auditing.
XAI provides a mechanism to assess whether the reasoning of a model is aligned with known physics, engineering constraints, or known causal mechanisms. 
Attribution tools such as saliency maps, and causal-discovery frameworks can reveal whether an AI system relies on physically meaningful features rather than spurious correlations. 
This is especially critical for rare but high-impact events, such as mechanical failures, or unstable aerodynamic regimes, where small precursors may carry disproportionate influence on safety. 
By interrogating internal representations, XAI enables scientists and engineers to verify that models use appropriate physical concepts, respect invariances, and behave consistently under perturbations. 
In this sense, explainability becomes an analogue to structural inspection: it clarifies \textit{why} a model behaves as it does, not merely \textit{how well} it behaves on historical tests (i.e., its accuracy).
If the model is able to satisfy known concepts, it is also more likely to generalize to unseen scenarios or observations because it might have learned the underlying mechanisms leading to a given outcome, rather than a spurious input-output map.
Generalizability is frequently tied to reduced ML-model ``hallucinations''. This is the case for recent hybrid physics-AI models~\cite{wang2025condensnet} and large foundation models with physical constraints.
Of course, one may argue that if we had a perfect model, we would not require such an understanding of its internal mechanisms for certification and auditing purposes, since the model would never fail. 
Yet, this is a rather utopian view of current state-of-the-art ML and more generally of the engineering and scientific world. And it does raise other and more profound questions of human and AI co-existence~\cite{superintelligence}.

A certification-oriented XAI framework would integrate three complementary pillars. 
First, physics-grounded explainability, whereby explanations are evaluated directly against conservation laws, stability properties, domain-specific concepts, or empirical scaling relations. 
Second, stress-testing via interpretable diagnostics, where distribution shifts, out-of-domain behaviour, and extreme-event responses are analysed through explanation-based stability metrics. 
These diagnostics can reveal failure modes long before they appear in conventional performance scores. 
Third, explanation-driven uncertainty quantification, where inconsistencies in attribution patterns serve as indicators of epistemic uncertainty. 
When these elements are combined, they provide an interpretable audit trail that regulators and domain experts can examine, analogous to verification and validation (V\&V) frameworks used in engineering certification.

Our view is that explainability is a core scientific instrument for certifying and auditing AI systems. 
It provides transparency about internal mechanisms, facilitates the identification of unphysical behaviours, and guides model redesign toward more robust and stable architectures. 
By coupling XAI methodologies with physics-based constraints, hybrid modelling, and rigorous V\&V principles, it becomes possible to envision a future where AI components in climate models, structural-dynamics solvers, or autonomous engineering platforms can be certified with the same rigour as traditional numerical models.
In medicine, explainability can be used as an AI colleague that explains its decision to its human counterpart, while the human takes the final decision more confidently, and with a transparent auditing trail~\cite{wei2025explainability}.

Such a framework would accelerate the safe integration of AI across science and engineering, enabling regulators, researchers, and industry partners to move from “trust by performance’’ to “trust by understanding.’’ 
This transition is essential for unlocking the full potential of AI-enabled discovery and ensuring that AI-driven predictions and decisions remain reliable, transparent, and scientifically grounded across the entire operational spectrum, especially in safety-critical applications, where failure means significant liabilities to the parties affected~\cite{VinuesaSirmacek2021NatMachIntell}.
\begin{figure}[H]
\centering
\includegraphics[width=\textwidth]{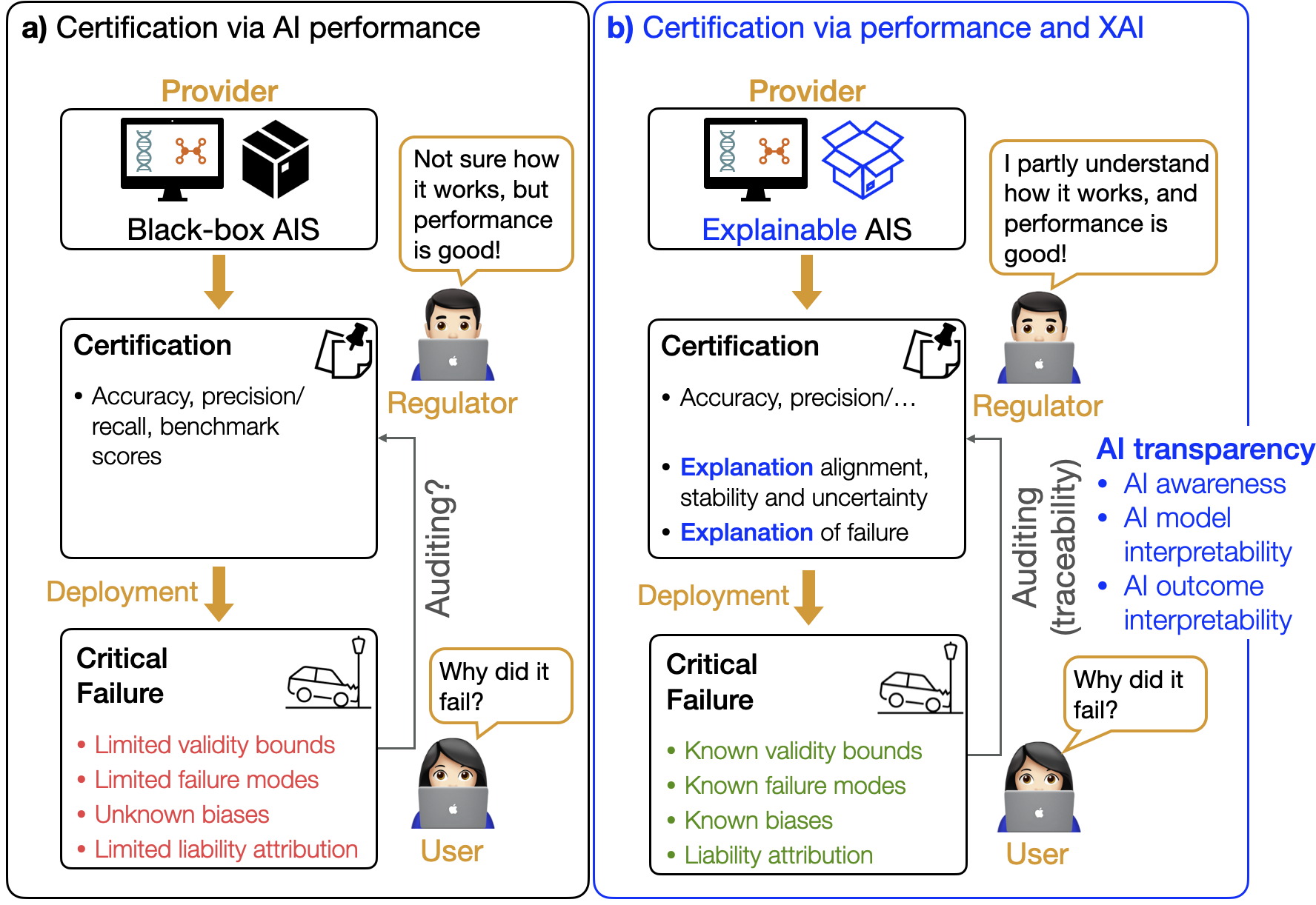}
\caption{{\bf Possible XAI-based framework for certification.} a) Shows a certification and auditing process obtained via the performance of the AI system (i.e., ``trust by performance''). This pathway to certification and auditing has critical issues, namely limited validity bounds, and understanding of failure modes, unknown biases, and limited liability attribution. b) Shows a certification and auditing process obtained through both performance and explainability (i.e., ``trust by understanding''), that addresses some of the issues of ``trust by performance''. Certification and auditing by performance and explainability allows for traceability and therefore auditing, and enables a better understanding of why a model may fail or not behave as expected, therefore giving the opportunity to address potential biases or unsafe behavior prior of it occurring.}
\label{fig:certification}
\end{figure}
%



\section*{Discussion and Outlook}

This Perspective has argued that explainable artificial intelligence provides a unifying framework for learning from machine-learned representations that already exceed human performance in specific tasks. Across discovery, optimization, and certification, explainability emerges as a tool that enables humans to extract causal structure, generalize beyond observed data and establish trust in high-stakes scientific and engineering applications. While these domains differ in objectives, acceptable risk and modes of validation, they share a common challenge: translating high-dimensional, machine-internal representations into forms of understanding that are meaningful and actionable for humans.

A first fundamental challenge concerns the intuitiveness of explanations. Not all explanations that are mathematically faithful are accessible to human experts. Attribution maps, latent variables, or causal graphs may be correct yet fail to align with the domain concepts, scales, or abstractions used in scientific reasoning. This mismatch poses a practical limit for certification and adoption: if explanations cannot be interpreted by regulators, engineers, or scientists, they cannot serve as a basis for accountability or decision making~\cite{Selbst2018ExplainableMachines}. 
Intuitiveness, however, should not be confused with simplicity. This concept reflects whether explanations can be composed, contextualized, and related to known concepts. This places an important constraint on XAI research: explanations must be evaluated not only for faithfulness to the model, but also for their usability by humans in specific scientific and engineering contexts.

A second unifying theme is the role of scientific understanding. Recent philosophical and methodological work has emphasized that understanding is achieved when a theory allows reliable reasoning across contexts without complete computation~\cite{Krenn2022ScientificUnderstandingAI}. In this sense, understanding is inseparable from abstraction: once the key mechanisms are identified, humans can anticipate system behavior, design interventions, or reason without re-solving the complete problem. This perspective offers a powerful lens for learning from the learners. If an ML model has internalized abstractions that enable strong generalization or efficient control, then extracting those abstractions allows humans to use part of the ML-based understanding. In other words, explainable AI can be a mechanism for transferring abstraction from machines to humans, and not simply justifying individual predictions.

Causality sits at the center of such a transfer. Across all three domains, the key distinction is not whether a model performs well, but whether it has captured relationships that are stable under intervention. In discovery, this determines whether the learned laws reflect generalizable physical mechanisms rather than dataset-specific correlations. In optimization, it governs whether the interventions improve the performance of the system robustly or merely exploit narrow regimes. In certification, it dictates whether failures can be anticipated, diagnosed and attributed. The growing integration of causal inference with representation learning and attribution methods points towards a new methodological framework in which the latent representations are not only compressed but causally structured. This causal organization is essential for scalability: without it, interpretability methods will not be able to handle the dimensionality and complexity of modern foundation models.

Another cross-cutting issue is the tension between performance and transparency. Highly expressive models often rely on representations that are efficient for machines but unintuitive for humans. In this article we propose that the explanations should operate at the level of mechanisms, invariances, and causal relations that matter for the task at hand. This perspective reframes explainability as a multi-level concept. A model may remain opaque internally while still yielding explanations that are sufficient for scientific insight, design decisions, or certification requirements. The appropriate level of explanation therefore depends on risk, context, and intended use, reinforcing the need for domain-specific standards rather than universal interpretability metrics.

Looking forward, several research directions appear particularly promising. First, there is a need for validation frameworks for explanations that assess causal faithfulness, stability under perturbations and consistency across scales. Second, scalable methods for causal discovery in learned latent spaces must be further developed, particularly for spatiotemporal and multi-physics systems. Third, human-AI interaction deserves greater attention: explanations are not static artifacts but part of an iterative dialogue in which humans refine questions, constraints, and abstractions. 

Progress in one domain will continue to inform the others. Tools developed for extracting governing equations can strengthen certification by revealing non-physical behavior; optimization studies can expose the limits of explanation-based control; and certification requirements can constrain model design in ways that ultimately improve scientific insight. Learning from the learners is therefore not a unidirectional process but a feedback loop: machines learn from data, humans learn from machines, and both co-evolve towards models that are understandable, reliable, and aligned with human values. In this emerging paradigm, agentic-AI systems that combine foundation models with explainability and causal reasoning can yield a completely new framework for accelerated and autonomous scientific discovery and design.

\section*{Acknowledgements}
SLB acknowledges support from the the Boeing Company and the National Science Foundation AI Institute in Dynamic Systems (grant number 2112085).

\bibliography{references}

\end{document}